\documentclass[conference]{IEEEtran}
\IEEEoverridecommandlockouts

\usepackage{cite}
\usepackage{amsmath,amssymb,amsfonts}
\usepackage{algorithmic}
\usepackage{graphicx}
\usepackage{textcomp}
\usepackage{xcolor}
\usepackage{paralist}
\usepackage{eurosym}
\usepackage{flushend}
\usepackage{url}
\usepackage[utf8]{inputenc}
\usepackage[colorlinks=true]{hyperref}
\usepackage{soul}
\usepackage{caption}
\usepackage[list=true]{subcaption}
\usepackage{booktabs}
\usepackage{tabu}
\usepackage{wrapfig}
\usepackage{arydshln}
\usepackage[normalem]{ulem}
\usepackage{multirow}
\usepackage{siunitx}

\usepackage[left=1.62cm,right=1.62cm,top=1.78cm]{geometry} 

\def\BibTeX{{\rm B\kern-.05em{\sc i\kern-.025em b}\kern-.08em
    T\kern-.1667em\lower.7ex\hbox{E}\kern-.125emX}}
    
\begin{document}

\title{Multimodal-to-Text Prompt Engineering in Large Language Models Using Feature Embeddings for GNSS Interference Characterization\\
\thanks{This work has been carried out within the DARCII project, funding code 50NA2401, sponsored by the German Federal Ministry for Economic Affairs and Climate Action (BMWK) and supported by the German Aerospace Center (DLR), the Bundesnetzagentur (BNetzA), and the Federal Agency for Cartography and Geodesy (BKG).}}

\author{\IEEEauthorblockN{Harshith Manjunath,
    Lucas Heublein,
    Tobias Feigl,
    Felix Ott}
  \IEEEauthorblockA{Fraunhofer Institute for Integrated Circuits IIS, Nürnberg, Germany}
  \IEEEauthorblockA{{\{harshith.manjunath, lucas.heublein, tobias.feigl, \underline{felix.ott}\}@iis.fraunhofer.de}}}

\maketitle

\begin{abstract}
Large language models (LLMs) are advanced AI systems applied across various domains, including NLP, information retrieval, and recommendation systems. Despite their adaptability and efficiency, LLMs have not been extensively explored for signal processing tasks, particularly in the domain of global navigation satellite system (GNSS) interference monitoring. GNSS interference monitoring is essential to ensure the reliability of vehicle localization on roads, a critical requirement for numerous applications. However, GNSS-based positioning is vulnerable to interference from jamming devices, which can compromise its accuracy. The primary objective is to identify, classify, and mitigate these interferences. Interpreting GNSS snapshots and the associated interferences presents significant challenges due to the inherent complexity, including multipath effects, diverse interference types, varying sensor characteristics, and satellite constellations. In this paper, we extract features from a large GNSS dataset and employ LLaVA to retrieve relevant information from an extensive knowledge base. We employ prompt engineering to interpret the interferences and environmental factors, and utilize t-SNE to analyze the feature embeddings. Our findings demonstrate that the proposed method is capable of visual and logical reasoning within the GNSS context. Furthermore, our pipeline outperforms state-of-the-art machine learning models in interference classification tasks.\\
Github: \href{https://gitlab.cc-asp.fraunhofer.de/darcy_gnss}{https://gitlab.cc-asp.fraunhofer.de/darcy\_gnss}
\end{abstract}
\begin{IEEEkeywords}
  Large Language Models, LLaVA, Multimodal-to-Text, Prompt Engineering, In-context Learning, Global Navigation Satellite System, Interference Characterization
\end{IEEEkeywords}

\section{Introduction}
\label{label_introduction}

Humans interact with the world through various channels, such as vision and language, with each channel offering distinct advantages for representing and communicating specific concepts. The aim is to develop a versatile assistant capable of effectively following multimodal vision-and-language instructions, aligning with human intent to execute a wide range of real-world tasks in dynamic environments~\cite{liu_li_wu}. Language-augmented foundational vision models~\cite{li_liu_li} have demonstrated strong performance in open-world visual understanding tasks, including classification~\cite{xiao_wu_xu}, object detection~\cite{zhong_yang_zhang}, semantic segmentation~\cite{li_weinberger_belongie}, as well as visual generation and editing~\cite{rombach_blattmann}. LLMs serve as a universal interface for a general-purpose assistant, enabling the explicit representation of various task instructions in language. The recent success of ChatGPT~\cite{openai} has exemplified the power of aligned LLMs in adhering to human instructions~\cite{liu_li_wu}. Pre-trained language models~\cite{he_liu_gao} are task-agnostic, extending to the learned hidden embedding space, where models such as recurrent neural networks or transformers are pre-trained on web-scale unlabeled text corpora for general tasks and subsequently fine-tuned for specific tasks. LLMs, characterized by their larger model size and enhanced language comprehension, are capable of \textit{in-context learning}~\cite{coda_forno_binz}, where they acquire new tasks from a small set of examples provided in the prompt during inference time~\cite{liu_li_li}. Through advanced application and augmentation techniques, LLMs can be deployed as AI agents -- artificial entities that perceive their environment, make decisions, and take actions. These agents often need to augment LLMs to access updated information from external knowledge bases and verify whether system actions yield the desired results~\cite{minaee_mikolov}.

\begin{figure}[!b]
    \centering
    \includegraphics[width=1.0\linewidth]{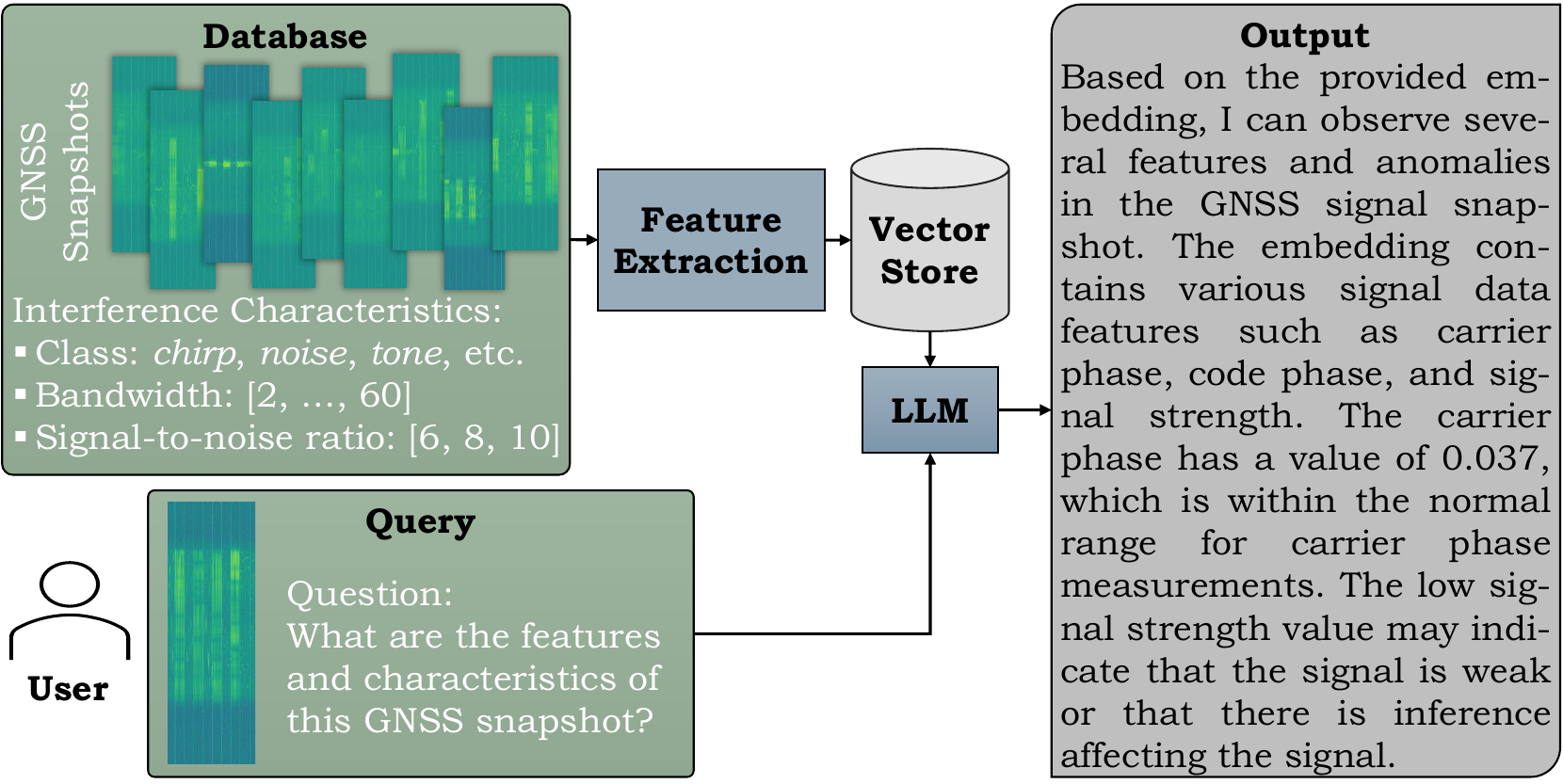}
    \caption{Based on feature embeddings extracted from GNSS snapshots, which include associated interference characteristics, a language model (LLM) generates a description in response to a contextual query provided by a user.}
    \label{figure_introduction_figure}
\end{figure}

Previous research has primarily focused on developing agents tailored to specific tasks and domains. However, in the context of GNSS interference monitoring~\cite{swinney_woods,ferre_fuente,li_huang_lang,xu_ying_li,ding_pham,gross_humphreys,heublein_feigl_crpa}, no studies have yet explored the use of language models for analyzing GNSS signals~\cite{esa_navisp,xiao_liu}, which have potential applications in crowdsourcing~\cite{raichur_ion_gnss}, aerospace systems, and toll collection management for highway trucking~\cite{ott_heublein_icl,ott_heublein_ecml}. The application of LLMs in these domains remains largely unexplored and is still at an early stage of research. The accuracy of GNSS receivers is significantly compromised by interference from jamming devices~\cite{crespillo_ruiz}, a problem that arises by the increasing availability of affordable and accessible jammers~\cite{merwe_franco}. Therefore, mitigating these interference signals is crucial, requiring the detection, classifation, and localization of interference source~\cite{brieger_ion_gnss}. However, analyzing GNSS signal and characterizing the interferences present substantial challenges due to the wide variability in interference bandwidths, signal-to-noise ratios, antenna characteristics, and environmental factors such as multipath effects~\cite{heublein_raichur_ion,brieger_ion_gnss,abraha_frisk}. The challenge lies in the multimodal vision-and-text approach to interference characterization, the transferability of techniques from image analysis to GNSS tasks~\cite{raichur_heublein}, and the analysis of snapshots through feature extraction. LLMs hold great promise for GNSS interference monitoring, thanks to their capability to process and interpret complex, multivariate data in real-time. They can accurately detect interference events while delivering explainable insights that strengthen system resilience and adaptability. Figure~\ref{figure_introduction_figure} illustrates our objective: to retrieve information from GNSS snapshots, characterize features using LLMs, and present a descriptive output to the end user, who may be a decision-maker, such as a non-expert interpreting the model's output in real-world operations~\cite{gianni_fuchs}. We evaluate the LLM output by examining prompt engineering in the context of GNSS interference monitoring systems.

\textbf{Contributions.} The primary objective of this work is to provide a detailed characterization of GNSS interferences using language models and prompt engineering. The novelty lies in adapting LLM capabilities to the specialized domain of GNSS interference monitoring by addressing key challenges; this includes adapting LLMs to process GNSS signal data, which fundamentally differs from text data. The key contributions of this research are as follows: (1) We introduce a method that leverages feature extraction and LLaVA~\cite{liu_li_wu} to generate descriptive outputs in response to user queries. (2) We present a comprehensive description of a GNSS dataset, focusing on its characteristics, including interference class, bandwidth, signal power, and multipath effects. (3) We explore prompt engineering by manually evaluating hundreds of query-output pairs with varying levels of detail. (4) We assess feature embeddings using t-SNE analysis. (5) We demonstrate that our proposed method outperforms traditional machine learning (ML) approaches in the task of interference classification.


\section{Related Work}
\label{label_related_work}

First, we provide a summary of related work that has evaluated prompt engineering for language models (see Section~\ref{label_rw_prompt_engineering}). Following this, we introduce the methods used for GNSS interference classification (see Section~\ref{label_rw_intf_class}).

\subsection{Prompt Engineering with Language Models}
\label{label_rw_prompt_engineering}

In the context of signal processing, Verma \& Pilanci~\cite{verma_pilanci} establish a comparison between classical Fourier transforms and learnable time-frequency representations for each intermediate activation signal within an LLM. Nguyen et al.~\cite{nguyen_nguyen} examine the vulnerabilities of LLMs in the context of 6G technology, particularly focusing on their susceptibility to malicious exploitation by analyzing known security weaknesses. Lin et al.~\cite{lin_qu_chen} highlight the potential of deploying LLMs at the 6G edge, emphasizing their ability to reduce long response times, high bandwidth costs, and data privacy violations. Yu et al.~\cite{yu_guo_sano} investigate the application of LLMs in diagnosing cardiac diseases and sleep apnea using ECG signals, by incorporating expert knowledge to guide the models beyond their inherent capabilities. Go et al.~\cite{go_han_chen} provide a comprehensive survey of prompt engineering, a technique that involves augmenting a large pre-trained model with task-specific prompts to adapt the model to new tasks. This approach allows for predictions based solely on the prompt, without requiring updates to the model parameters, which aligns with our proposed method. We extract features from a large database without fine-tuning the language model. However, this method requires complex reasoning to identify model errors, hypothesize about gaps or misleading aspects in the current prompt, and communicate the task with clarity. Its potential is limited due to lack of sufficient guidance for complex reasoning, as well as the need for detailed descriptions, context specification, and a structured reasoning framework~\cite{ye_axmed}. Nevertheless, prompt engineering and language models have yet to be applied to GNSS interference monitoring, which we address in the subsequent section.

\subsection{GNSS Interference Classification}
\label{label_rw_intf_class}

Several recent methods have focused on the classification of GNSS interferences. For instance, Swinney et al.~\cite{swinney_woods} explored the use of jamming signal power spectral density, spectrogram, raw constellation, and histogram signal representations as images to apply transfer learning from the imagery domain. Ferre et al.~\cite{ferre_fuente} employed support vector machines and convolutional neural networks to classify jammer types in GNSS signals, whereas Li et al.~\cite{li_huang_lang} and Xu et al.~\cite{xu_ying_li} adopted a twin SVM-based approach. Ding et al.~\cite{ding_pham} leveraged ML models in a single static (line-of-sight) propagation environment; in contrast, we extend this work by considering multipath environments. Gross et al.~\cite{gross_humphreys} utilized a maximum likelihood method to ascertain whether a synthetic signal is compromised by multipath or jamming. Heublein et al.~\cite{heublein_raichur_ion} highlighted the challenges associated with data discrepancies in the GNSS context. Few-shot learning~\cite{ott_heublein_icl} has been applied in the GNSS context to integrate new classes into a support set, aiming for a more continuous representation between positive and negative interference pairs. Raichur et al.~\cite{raichur_heublein} adapted to novel interference classes through continual learning. Brieger et al.~\cite{brieger_ion_gnss} incorporated both spatial and temporal relationships between samples by utilizing a joint loss function and a late fusion technique. Furthermore, Raichur et al.~\cite{raichur_ion_gnss} introduced a crowdsourcing approach that leverates smartphone-based features to localize the source of detected interference. For a comprehensive overview of recent adaptive GNSS applications, refer to Ott et al.~\cite{ott_heublein_ecml}. Gaikwad et al.~\cite{gaikwad_heublein} proposed a federated learning approach using few-shot learning and aggregation of the model weights on a global server by introducing a dynamic early stopping method to balance out-of-distribution classes based on representation learning, specifically utilizing the maximum mean discrepancy of feature embeddings between local and global models. Heublein et al.~\cite{heublein_feigl_posnav} proposed an ML approach that achieves high generalization in classifying interference through orchestrated monitoring stations deployed along highways. The presented semi-supervised approach is coupled with an uncertainty-based voting mechanism by combining Monte Carlo and Deep Ensembles that effectively minimizes the requirement for labeled training samples to less than 5\% of the dataset while improving adaptability across varying environments.

\begin{figure}[!t]
    \centering
    \includegraphics[width=1.0\linewidth]{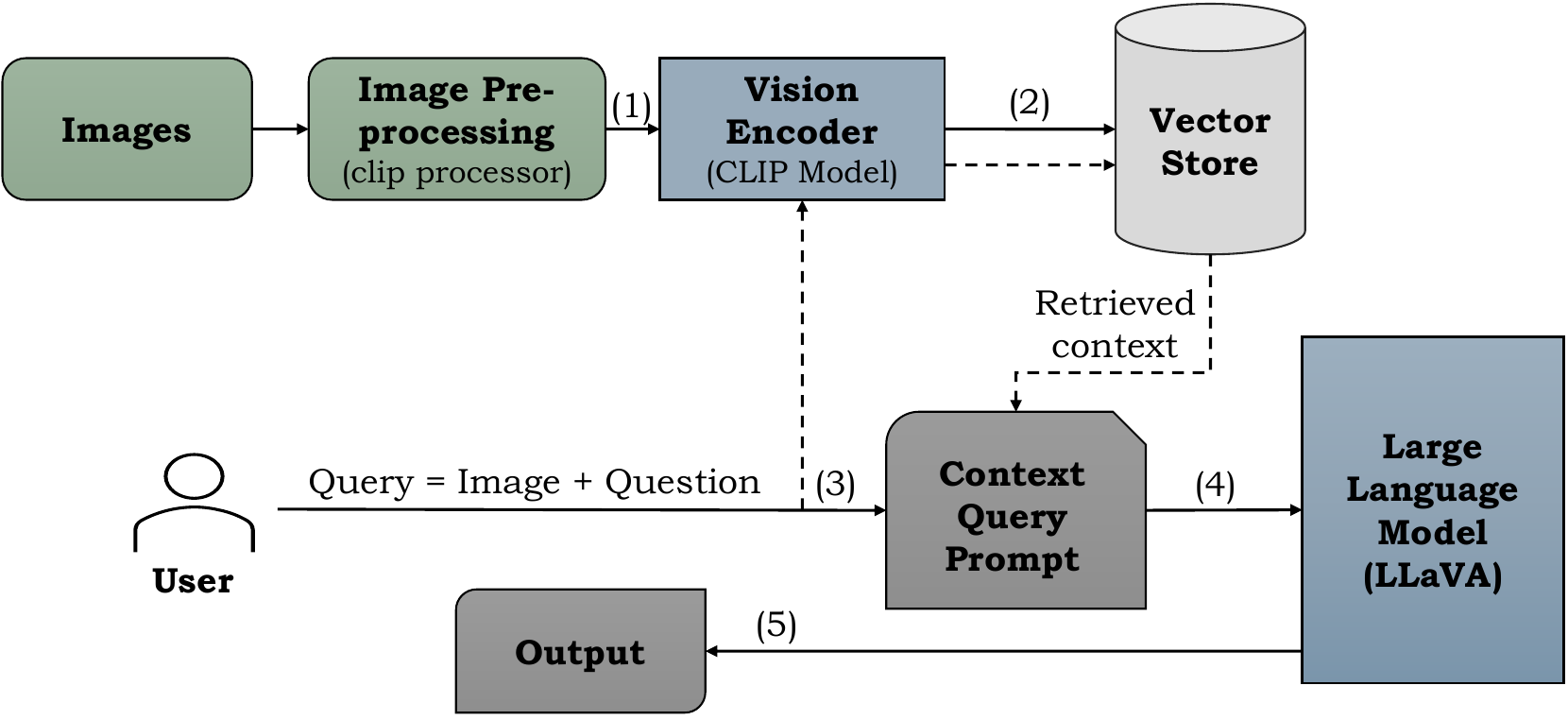}
    \caption{\textbf{Method overview}. GNSS snapshots are processed (1), and the resulting image embeddings are stored as vectors (2). When the user submits a context query prompt (3), the LLM (4) provides the corresponding output (5).}
    \label{figure_method_overview}
\end{figure}

\begin{figure}[!t]
    \centering
    \includegraphics[width=1.0\linewidth]{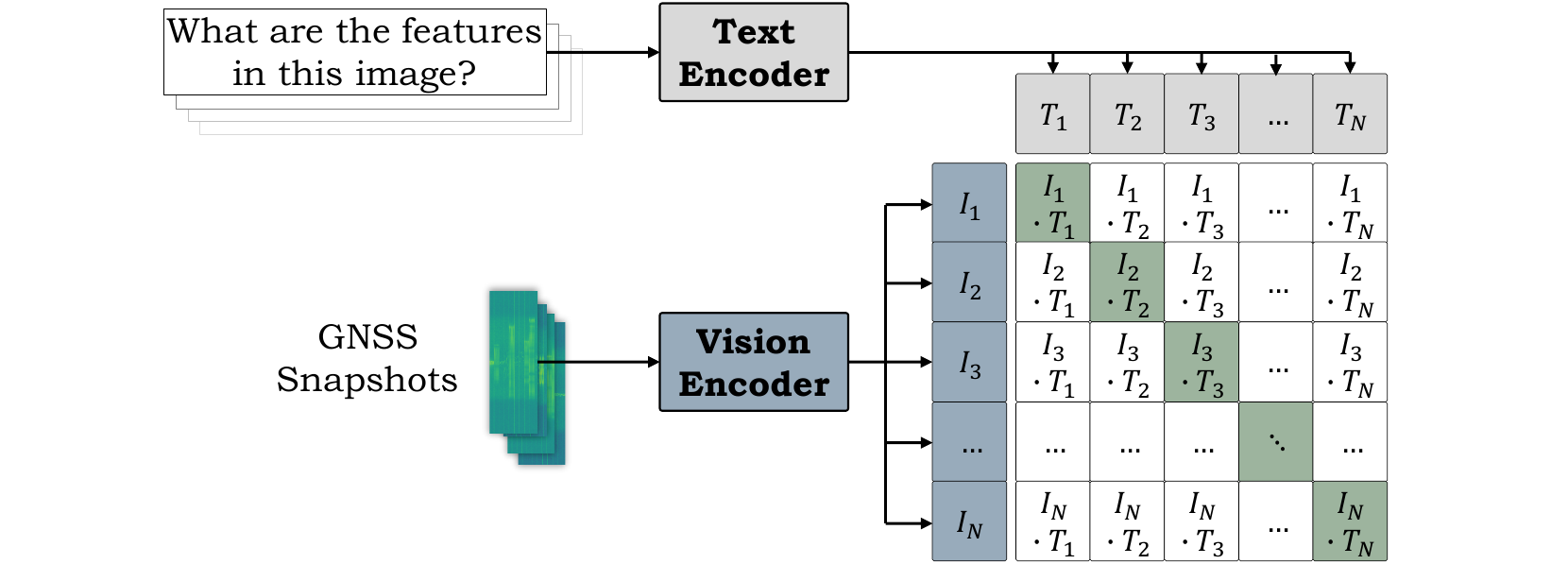}
    \caption{Overview of CLIP of an image-text pair input~\cite{radford_kim}.}
    \label{figure_clip_vit}
\end{figure}

\section{Methodology}
\label{label_method}

First, we provide an overview of the method. Following that, we introduce the embedding and LLM model. Lastly, we present the specifics of our prompt engineering approach.

\textbf{Method Overview.} Figure~\ref{figure_method_overview} presents a detailed overview of the proposed pipeline. The input image has dimensions of $1,024 \times 34$, and the dataset comprises a total of 42,592 GNSS snapshots. Initially, the images are processed using the CLIP (contrastive language-image pre-training) visual encoder ViT-L/14~\cite{radford_kim}. The features extracted by the CLIP model are stored as embeddings in a vector store. The process continues when the user submits a query, which includes an image (a GNSS snapshot) and a related question. Using a context query prompt, we instruct the LLM on its task. The LLM employed is based on LLaVA~\cite{liu_li_wu}, and the output is constrained to a maximum of 500 tokens.

\textbf{Vision Encoder.} We employ the CLIP\footnote{CLIP encoder: \url{https://huggingface.co/zer0int/CLIP-GmP-ViT-L-14}} ViT-L/14 vision encoder (refer to Figure~\ref{figure_clip_vit}), as proposed by Radford et al.~\cite{radford_kim}. CLIP is a neural network trained on a diverse set of (image, text) pairs. It is designed to predict the most relevant text snippet for a given image based on natural language instructions, without direct task-specific optimization. This approach facilitates zero-shot transfer of the model to downstream tasks.

\begin{figure}[!t]
    \centering
    \includegraphics[width=1.0\linewidth]{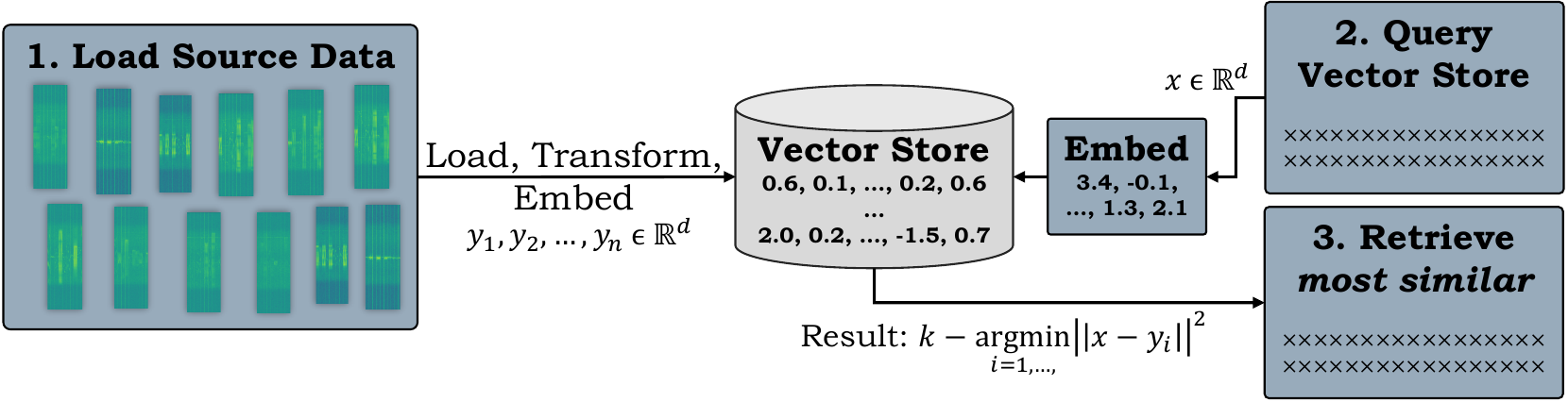}
    \caption{Overview of the vector store~\cite{douze_guzhva}.}
    \label{figure_vector_store}
\end{figure}

\textbf{Vector Store.} A prevalent method for storing and searching unstructured data involves embedding the data, storing the resulting embedding vectors, and then embedding the unstructured query at query time to retrieve the vectors that are most similar to the embedded query. A vector store manages the storage of embedded data and executes vector searches (refer to Figure~\ref{figure_vector_store}). In this work, we utilize the Facebook AI similarity search (FAISS) vector database~\cite{douze_guzhva} to store embeddings of size $(1, 512)$.

\begin{figure}[!t]
    \centering
    \includegraphics[width=1.0\linewidth]{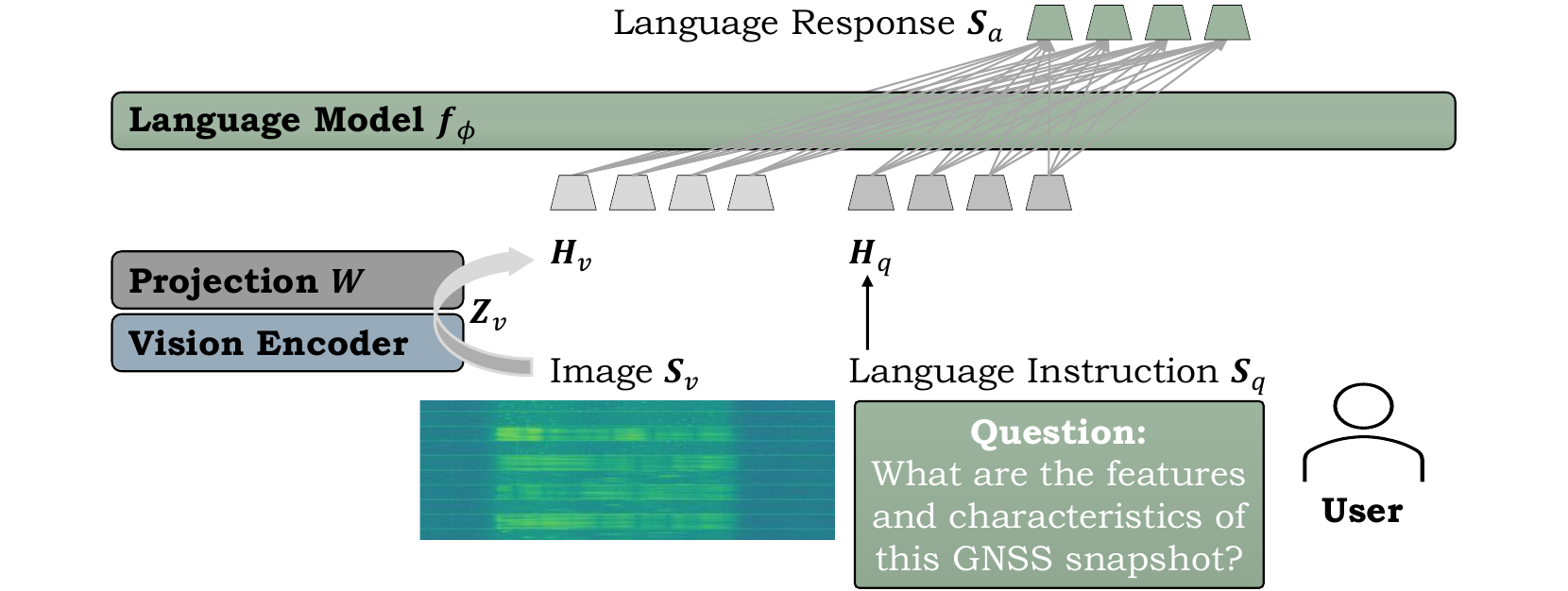}
    \caption{Network architecture of LLaVA~\cite{liu_li_wu}.}
    \label{figure_llava}
\end{figure}

\newcommand\y{0.0569}
\begin{figure*}[!t]\captionsetup[subfigure]{font=scriptsize}
    \centering
	\begin{minipage}[t]{\y\linewidth}
        \centering
    	\includegraphics[trim=192 60 176 50, clip, width=1.0\linewidth]{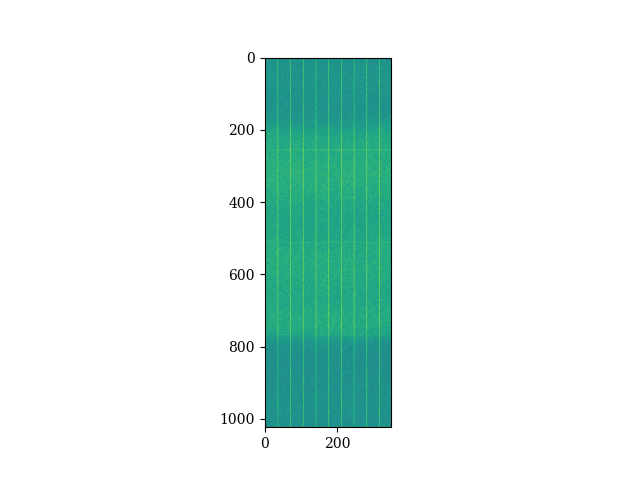}
    	\subcaption{None.}
    	\label{figure_exmp_samples1}
    \end{minipage}
    \hfill
	\begin{minipage}[t]{\y\linewidth}
        \centering
    	\includegraphics[trim=192 60 176 50, clip, width=1.0\linewidth]{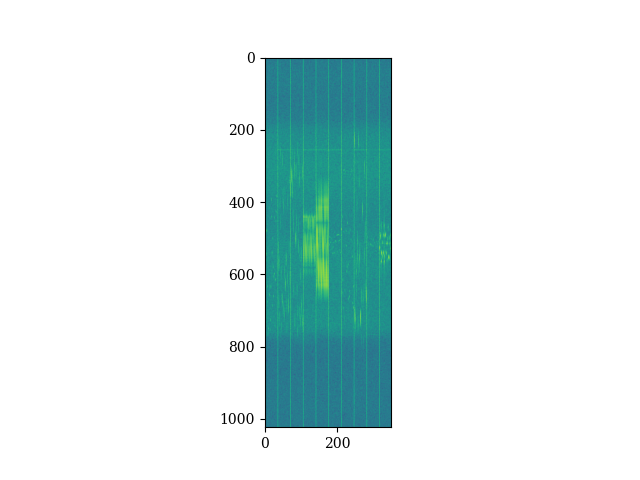}
    	\subcaption{Chirp.}
    	\label{figure_exmp_samples2}
    \end{minipage}
    \hfill
	\begin{minipage}[t]{\y\linewidth}
        \centering
    	\includegraphics[trim=192 60 176 50, clip, width=1.0\linewidth]{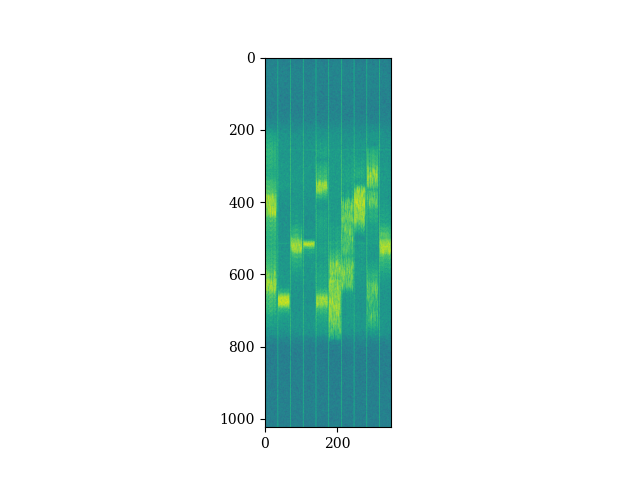}
    	\subcaption{Freq-Hopper.}
    	\label{figure_exmp_samples3}
    \end{minipage}
    \hfill
	\begin{minipage}[t]{\y\linewidth}
        \centering
    	\includegraphics[trim=192 60 176 50, clip, width=1.0\linewidth]{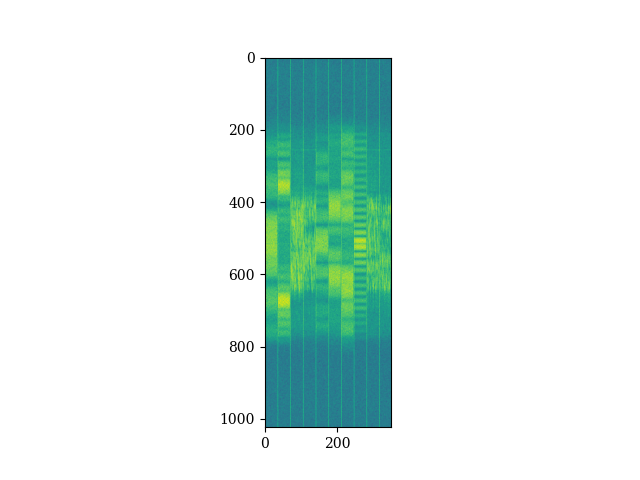}
    	\subcaption{Modulated.}
    	\label{figure_exmp_samples4}
    \end{minipage}
    \hfill
	\begin{minipage}[t]{\y\linewidth}
        \centering
    	\includegraphics[trim=192 60 176 50, clip, width=1.0\linewidth]{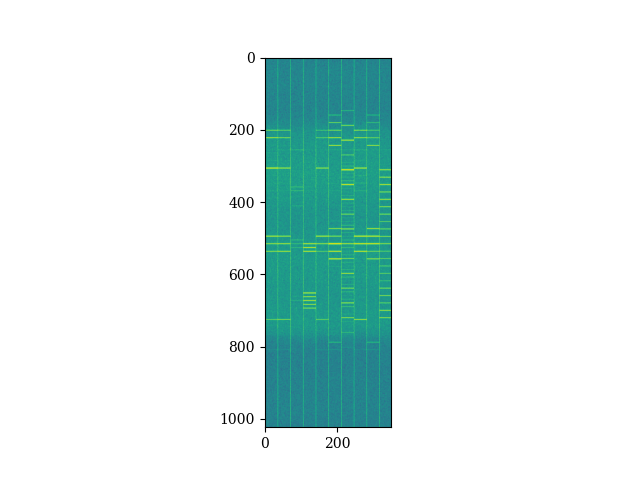}
    	\subcaption{Multitone.}
    	\label{figure_exmp_samples5}
    \end{minipage}
    \hfill
	\begin{minipage}[t]{\y\linewidth}
        \centering
    	\includegraphics[trim=192 60 176 50, clip, width=1.0\linewidth]{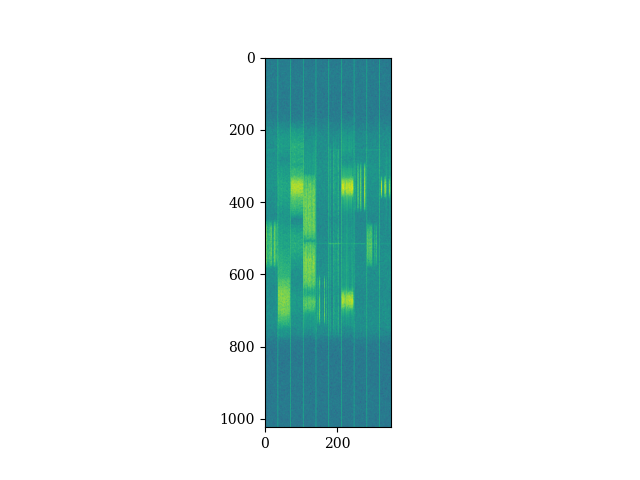}
    	\subcaption{Pulsed.}
    	\label{figure_exmp_samples6}
    \end{minipage}
    \hfill
	\begin{minipage}[t]{\y\linewidth}
        \centering
    	\includegraphics[trim=192 60 176 50, clip, width=1.0\linewidth]{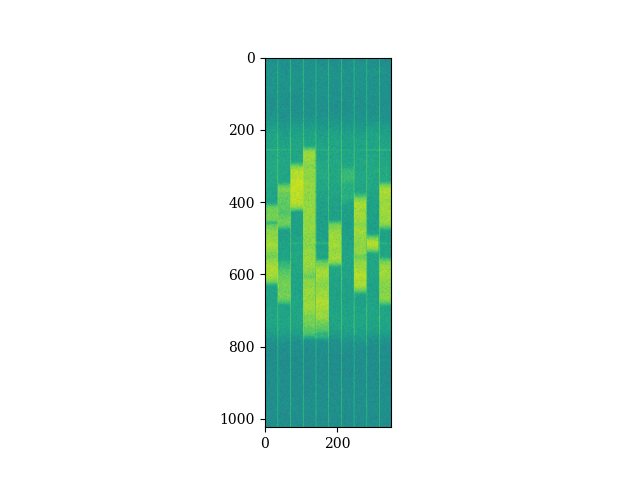}
    	\subcaption{Noise.}
    	\label{figure_exmp_samples7}
    \end{minipage}
    \hfill
	\begin{minipage}[t]{\y\linewidth}
        \centering
    	\includegraphics[trim=192 60 176 50, clip, width=1.0\linewidth]{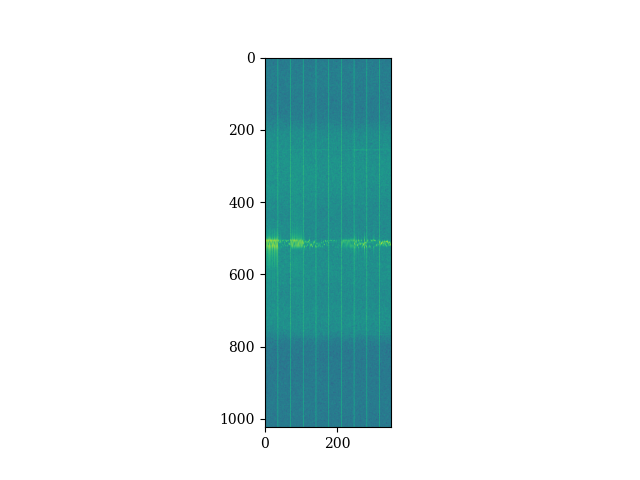}
    	\subcaption{Chirp, BW 2.}
    	\label{figure_exmp_samples8}
    \end{minipage}
    \hfill
	\begin{minipage}[t]{\y\linewidth}
        \centering
    	\includegraphics[trim=192 60 176 50, clip, width=1.0\linewidth]{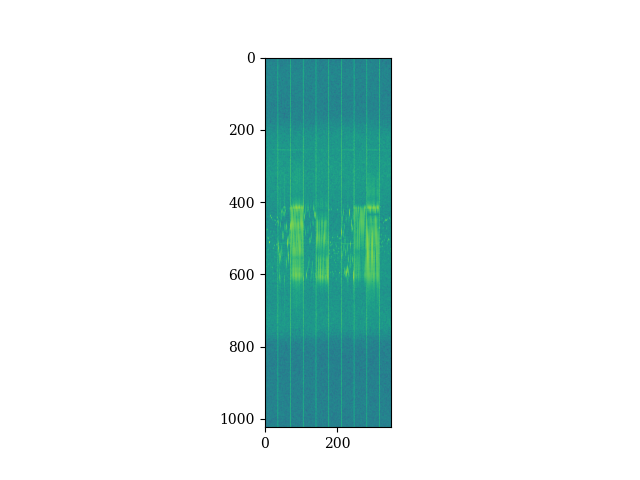}
    	\subcaption{Chirp, BW 20.}
    	\label{figure_exmp_samples9}
    \end{minipage}
    \hfill
	\begin{minipage}[t]{\y\linewidth}
        \centering
    	\includegraphics[trim=192 60 176 50, clip, width=1.0\linewidth]{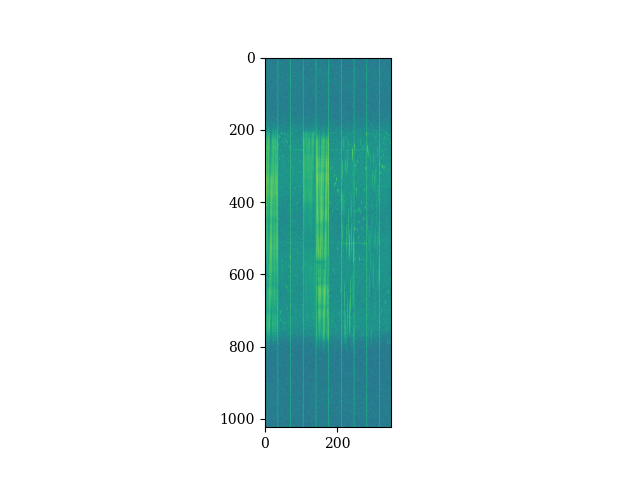}
    	\subcaption{Chirp, BW 60.}
    	\label{figure_exmp_samples10}
    \end{minipage}
    \hfill
	\begin{minipage}[t]{\y\linewidth}
        \centering
    	\includegraphics[trim=192 60 176 50, clip, width=1.0\linewidth]{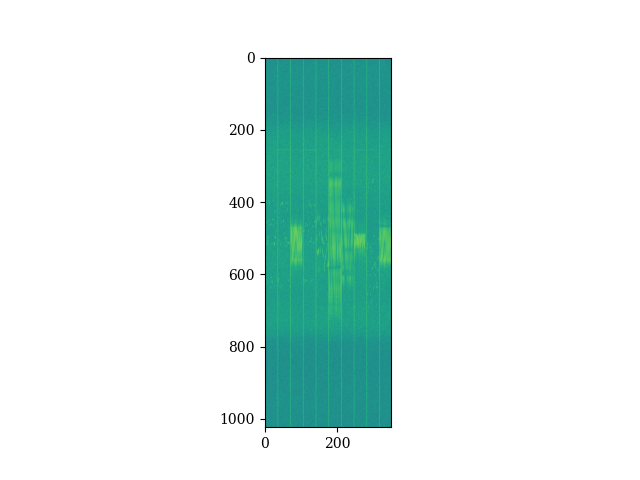}
    	\subcaption{Chirp, power -10.}
    	\label{figure_exmp_samples11}
    \end{minipage}
    \hfill
	\begin{minipage}[t]{\y\linewidth}
        \centering
    	\includegraphics[trim=192 60 176 50, clip, width=1.0\linewidth]{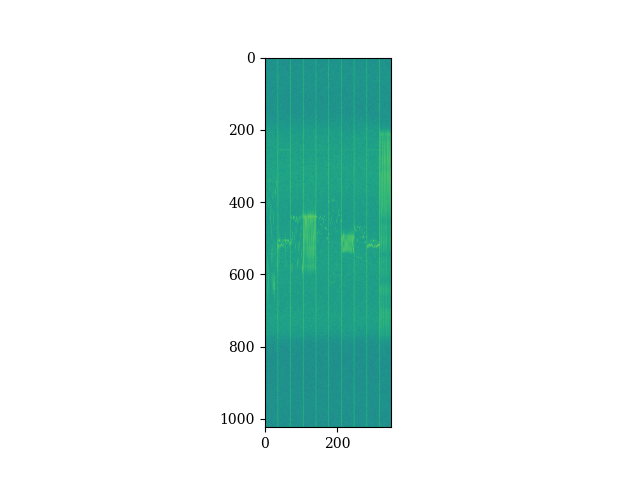}
    	\subcaption{Chirp, power 4.}
    	\label{figure_exmp_samples12}
    \end{minipage}
    \hfill
	\begin{minipage}[t]{\y\linewidth}
        \centering
    	\includegraphics[trim=192 60 176 50, clip, width=1.0\linewidth]{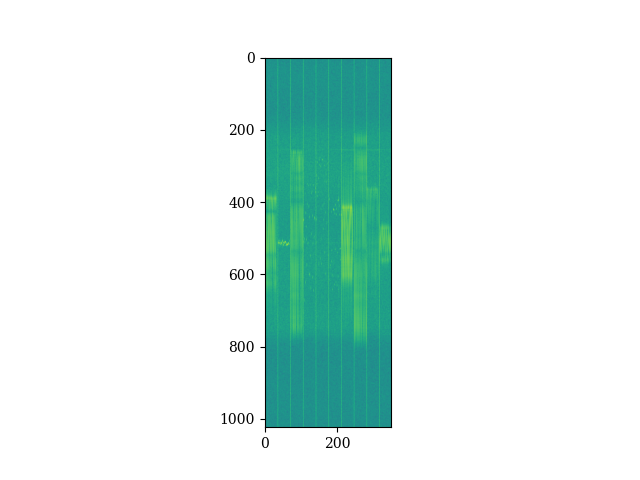}
    	\subcaption{Chirp, power 10.}
    	\label{figure_exmp_samples13}
    \end{minipage}
    \hfill
	\begin{minipage}[t]{\y\linewidth}
        \centering
    	\includegraphics[trim=192 60 176 50, clip, width=1.0\linewidth]{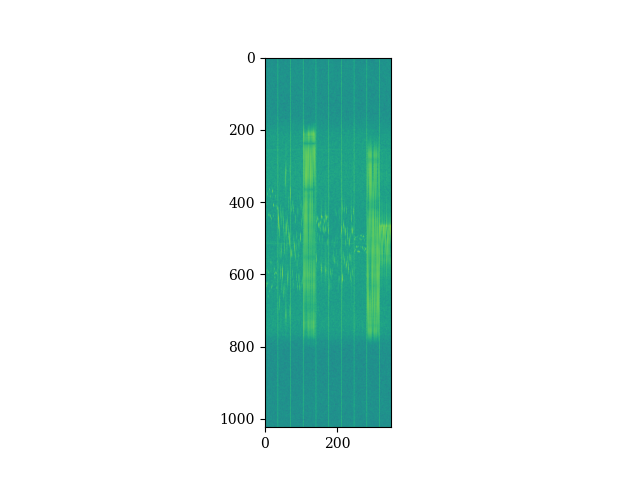}
    	\subcaption{Chirp, scenario 1.}
    	\label{figure_exmp_samples14}
    \end{minipage}
    \hfill
	\begin{minipage}[t]{\y\linewidth}
        \centering
    	\includegraphics[trim=192 60 176 50, clip, width=1.0\linewidth]{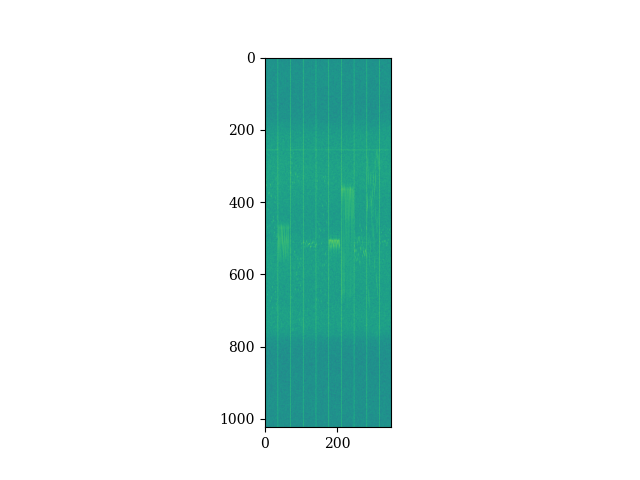}
    	\subcaption{Chirp, scenario 7.}
    	\label{figure_exmp_samples15}
    \end{minipage}
    \hfill
	\begin{minipage}[t]{\y\linewidth}
        \centering
    	\includegraphics[trim=192 60 176 50, clip, width=1.0\linewidth]{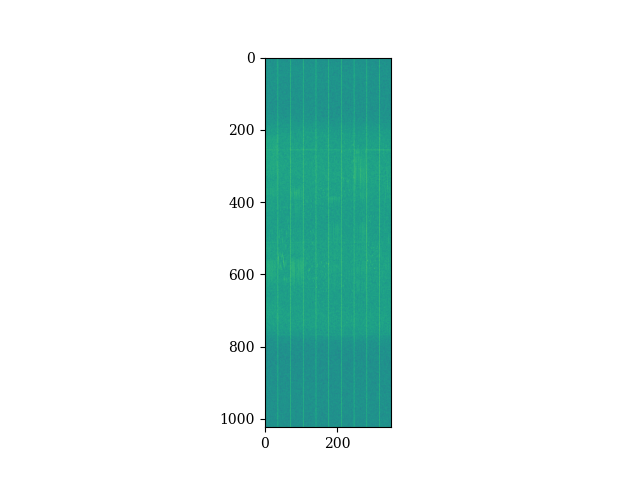}
    	\subcaption{Chirp, scenario 8.}
    	\label{figure_exmp_samples16}
    \end{minipage}
    \caption{Exemplary snapshot samples (concatenation of 10 samples) of the non-interference class (a) and all six interference types (b to g), a signal with chirp interference with different bandwidths (BW) (h to j) and signal powers (k to m), and a chirp interference from the scenario 1 (open environment), scenario 7, and scenario 8. Figures from Heublein et al.~\cite{heublein_feigl_crpa}.}
    \label{figure_exmp_samples}
\end{figure*}

\textbf{Language Model.} The objective is to effectively harness the capabilities of both the pre-trained LLM and the visual model. We employ the LLaVA\footnote{LLaVA model: \url{https://huggingface.co/llava-hf/llava-v1.6-mistral-7b-hf}} model~\cite{liu_li_wu}, which is based on Vicuna~\cite{chiang_li_lin}, as the language model $f_\phi(\cdot)$, parameterized by $\phi$. Although LLaVA-NeXT~\cite{liu_li_li} is too large for our current hardware configuration\footnote{All experiments were conducted using Nvidia Tesla V100-SXM2 GPUs with 32 GB VRAM, alongside Core Xeon CPUs and 192 GB RAM.}, it remains a potential architecture for future work. Figure~\ref{figure_llava} illustrates the network architecture, where a GNSS snapshot is used as input, and user queries serve as language instructions. Given an input image snapshot $\mathbf{S}_v$, the visual feature $\mathbf{Z}_v = g(\mathbf{S}_v)$ is extracted using the pre-trained CLIP visual encoder ViT-L/14~\cite{radford_kim}. A linear layer maps the image features into the word embedding space; specifically, the visual features $\mathbf{Z}_v$ are projected via the matrix $\mathbf{W}$ into language embedding tokens $\mathbf{H}_v$ by $\mathbf{H}_v = \mathbf{W} \cdot \mathbf{Z}_v$~\cite{liu_li_wu}. During the training phase, multi-turn conversation data $(\mathbf{S}_q^1, \mathbf{S}_a^1, \ldots, \mathbf{S}_q^T, \mathbf{S}_a^T)$ is generated for each image $\mathbf{S}_v$, where $T$ represents the total number of turns. Following the approach used in LLaVA, we conduct instruction-tuning of the LLM on the prediction tokens using an auto-regressive training objective: For a sequence of length $L$, the probability of the target answers $\mathbf{S}_a$ are computed by
\begin{equation}
\label{equ_instruct}
    p(\mathbf{S}_a | \mathbf{S}_v, \mathbf{S}_{\text{instruct}}) = \prod_{i=1}^{L} p_{\theta}(s_i | \mathbf{S}_v, \mathbf{S}_{\text{instruct},<i}, \mathbf{S}_{a,<i}),
\end{equation}
where $\mathbf{S}_{\text{instruct},<i}$ are the instruction tokens and $\mathbf{S}_{a,<i}$ are the answer tokens in all turns before the current prediction token $s_i$~\cite{liu_li_wu}.

\textbf{Prompt Engineering.} Visual LLMs have the potential to provide a more comprehensive understanding of multimodal data by combining textual and visual cues, thereby generating outputs that more closely align with human-like reasoning and perception. This fusion process allows the model to capture interdependencies and interactions between textual and visual elements, leading to more accurate and contextually grounded outputs. For instance, when additional textual information about GNSS data is provided by the user, the model can better integrate this input. A well-designed fusion module is crucial for capturing interactions and relationships between modalities, avoiding semantic mismatches, and mitigating biases~\cite{go_han_chen}. \textit{Task instruction prompting}~\cite{efrat_levy} involves using carefully designed prompts that provide explicit task-related instructions to guide the model’s behavior; defined as $x_{\text{input}} = \mathcal{H}(\mathbf{S}_v, t)$, where $\mathcal{H}$ represents the task instruction function, taking the image $\mathbf{S}_v$ and text $t$ as inputs to produce a modified input representation $x_{\text{input}}$. \textit{In-context learning}~\cite{wei_wang_schuurmans,dong_li_dai} is a method where the model is presented with a sequence of related examples or prompts, enabling it to learn and generalize from the provided context; defined as $x_{\text{input}} = \mathcal{H}(\mathcal{C}, \mathbf{S}_v, t)$, where $\mathcal{H}$ integrates the given context $\mathcal{C}$ with the image $\mathbf{S}_v$ and text $t$ inputs~\cite{go_han_chen}. In our approach, we utilize in-context learning by incorporating the retrieved context from the vector store into the user query (refer to Figure~\ref{figure_method_overview}).

\section{Dataset}
\label{label_dataset}

\newcommand\x{0.24}
\begin{figure}[!t]
    \centering
	\begin{minipage}[t]{\x\linewidth}
       \centering
    	\includegraphics[trim=200 120 20 0, clip, width=1.0\linewidth]{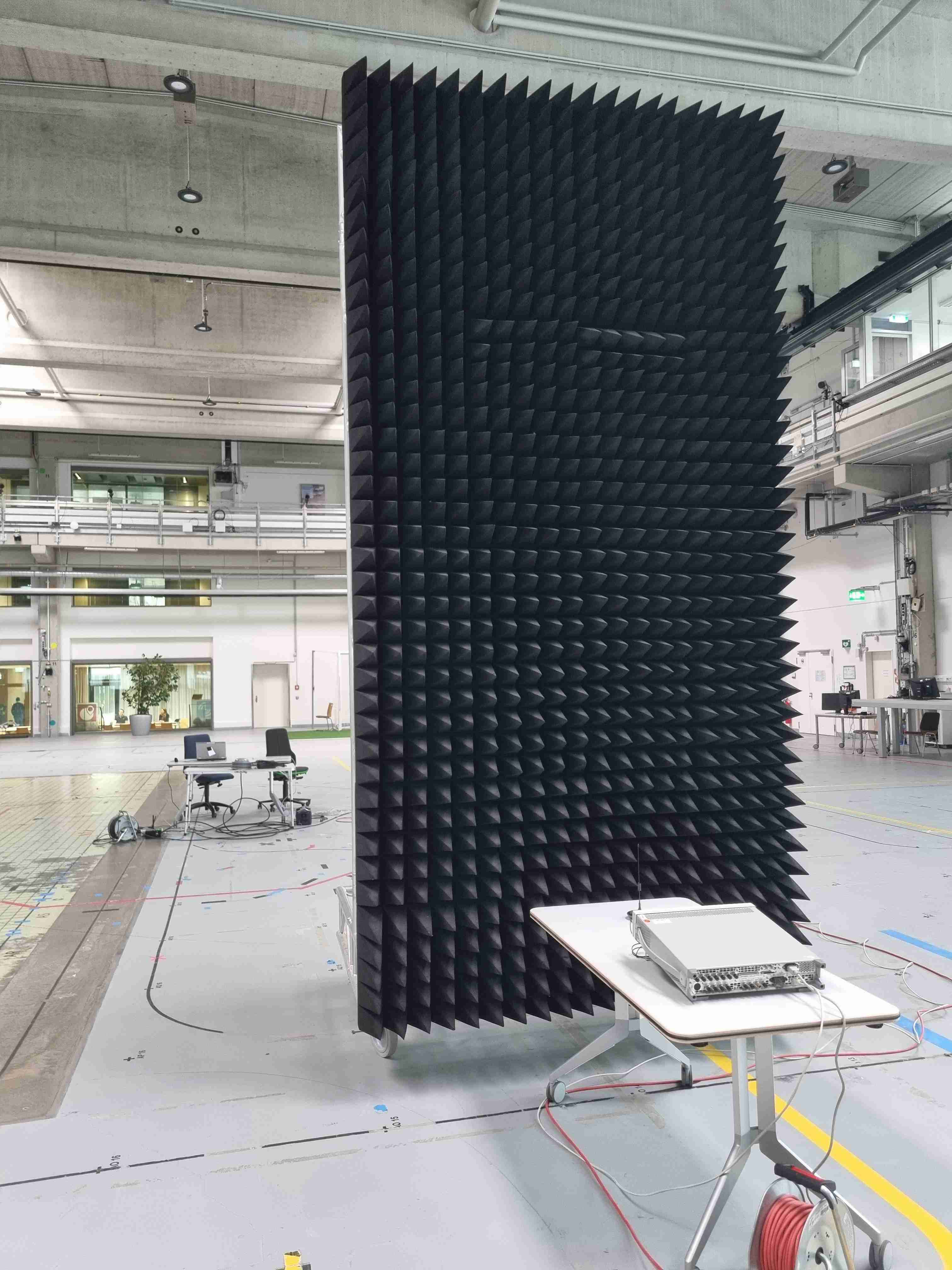}
    	\subcaption{Scenario 2.}
    	\label{figure_multi_path2}
    \end{minipage}
    \hfill
	\begin{minipage}[t]{\x\linewidth}
        \centering
    	\includegraphics[trim=110 60 110 60, clip, width=1.0\linewidth]{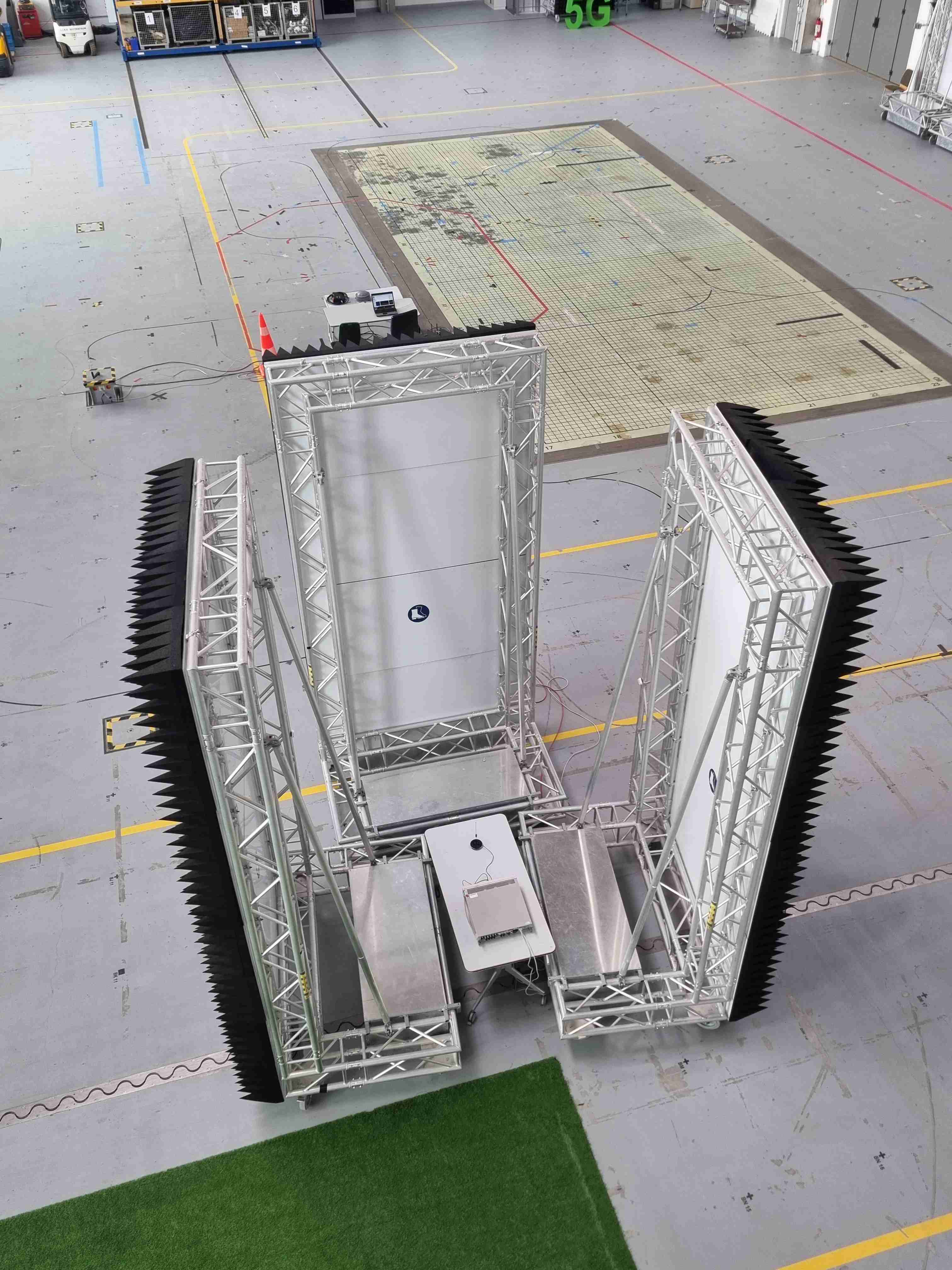}
    	\subcaption{Scenario 5.}
    	\label{figure_multi_path5}
    \end{minipage}
    \hfill
	\begin{minipage}[t]{\x\linewidth}
        \centering
    	\includegraphics[trim=190 40 190 80, clip, width=1.0\linewidth]{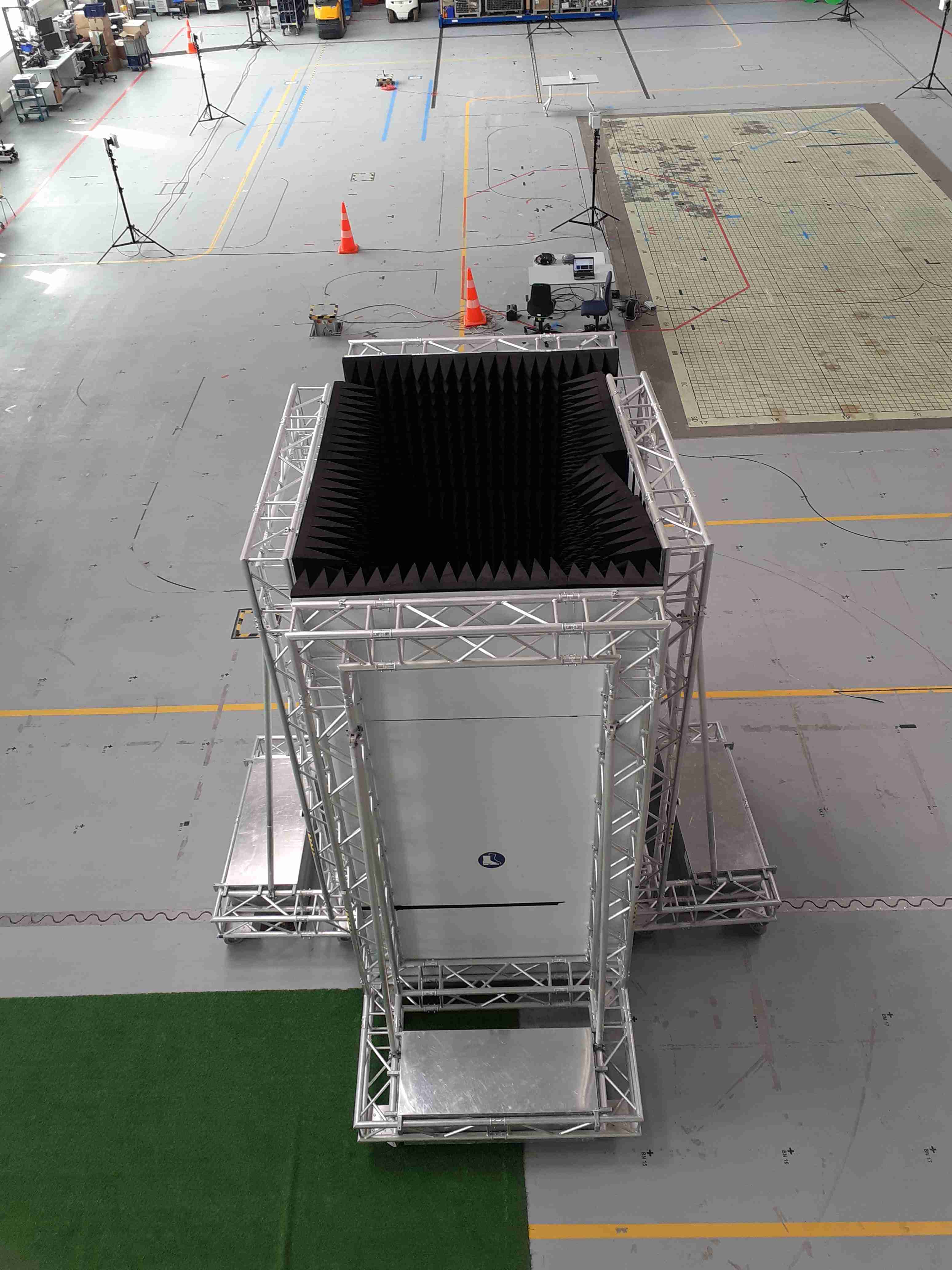}
    	\subcaption{Scenario 7.}
    	\label{figure_multi_path7}
    \end{minipage}
    \hfill
	\begin{minipage}[t]{\x\linewidth}
        \centering
    	\includegraphics[trim=24 0 24 0, clip, width=1.0\linewidth]{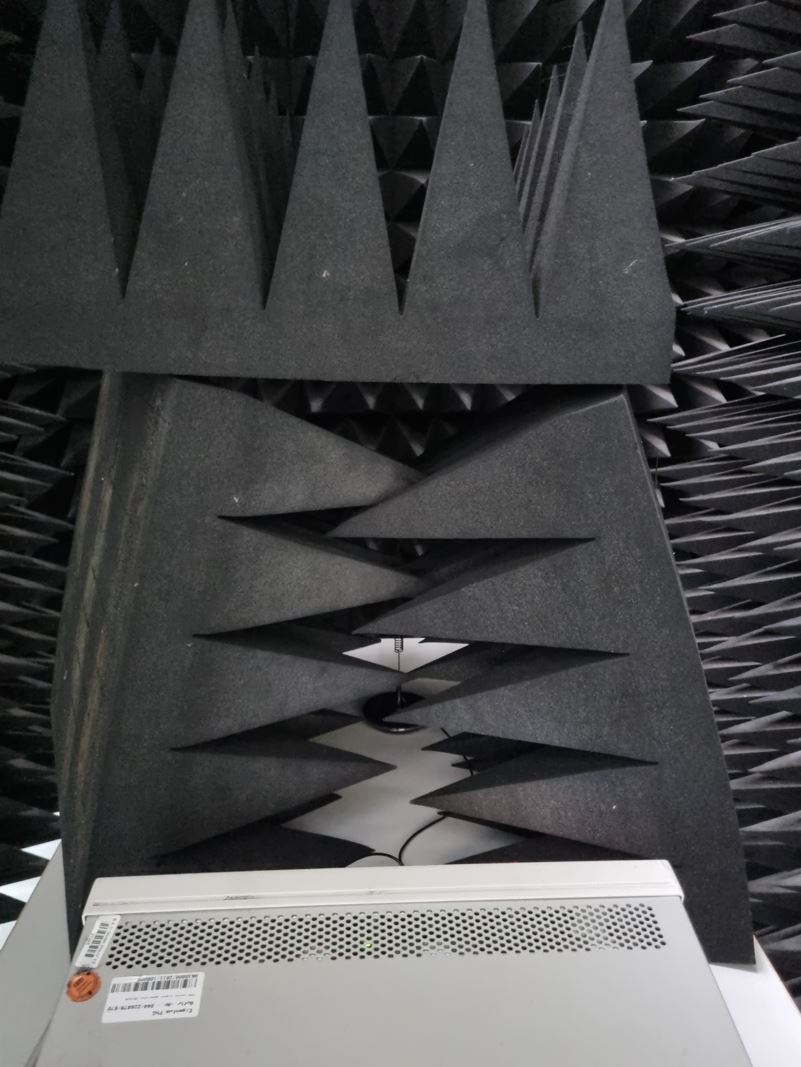}
    	\subcaption{Scenario 8.}
    	\label{figure_multi_path8}
    \end{minipage}
    \caption{Overview of different multipath scenarios where large black absorber walls are placed between and around the signal generator and the antenna (from Heublein et al.~\cite{heublein_feigl_crpa}).}
    \label{figure_multi_path}
\end{figure}

We employ a GNSS snapshot-based dataset as introduced by Ott et al.~\cite{heublein_feigl_crpa}. The primary objective is to develop ML models that demonstrate robustness against various types of jammers, interference characteristics, antenna variations, and environmental changes. Data collection takes place within the Fraunhofer IIS L.I.N.K.~center, a controlled environment measuring $1,320\,\textit{m}^2$, which facilitates the incorporation of multipath effects. A receiver antenna is positioned at one end of the hall, while an MXG vector signal generator is placed at the opposite end. The signal generator is capable of producing high-quality radio frequency signals across a wide range of frequencies. The signals received by the antenna are recorded as snapshots that include various interferences introduced by the signal generator. A comprehensive dataset is gathered under different configurations, including scenarios within an empty environment and setups where absorber walls are placed between the antenna and the generator (as shown in Figure~\ref{figure_multi_path}). The antenna captures snapshots at a frequency of $100\,\text{MHz}$ with a duration of $10\,\mu s$. Each snapshot consists of 1,024 channels, resulting in dimensions of $1,024 \times 34$. Figure~\ref{figure_exmp_samples} presents a selection of snapshots, with each image showcasing 10 randomly selected samples. Six distinct types of interference are introduced. Furthermore, the jammer types are characterized py parameters such as bandwidth (BW), ranging from 0.1 to 60, and signal power levels, ranging from -10 to 10. It is noted that increased multipath effects correspond to a reduction in interference intensity. The dataset contains a total of 42,592, of which 576 are free from any interferences~\cite{heublein_feigl_crpa}.

\section{Evaluation}
\label{label_evaluation}

\begin{figure}[!t]
    \centering
    \includegraphics[width=1.0\linewidth]{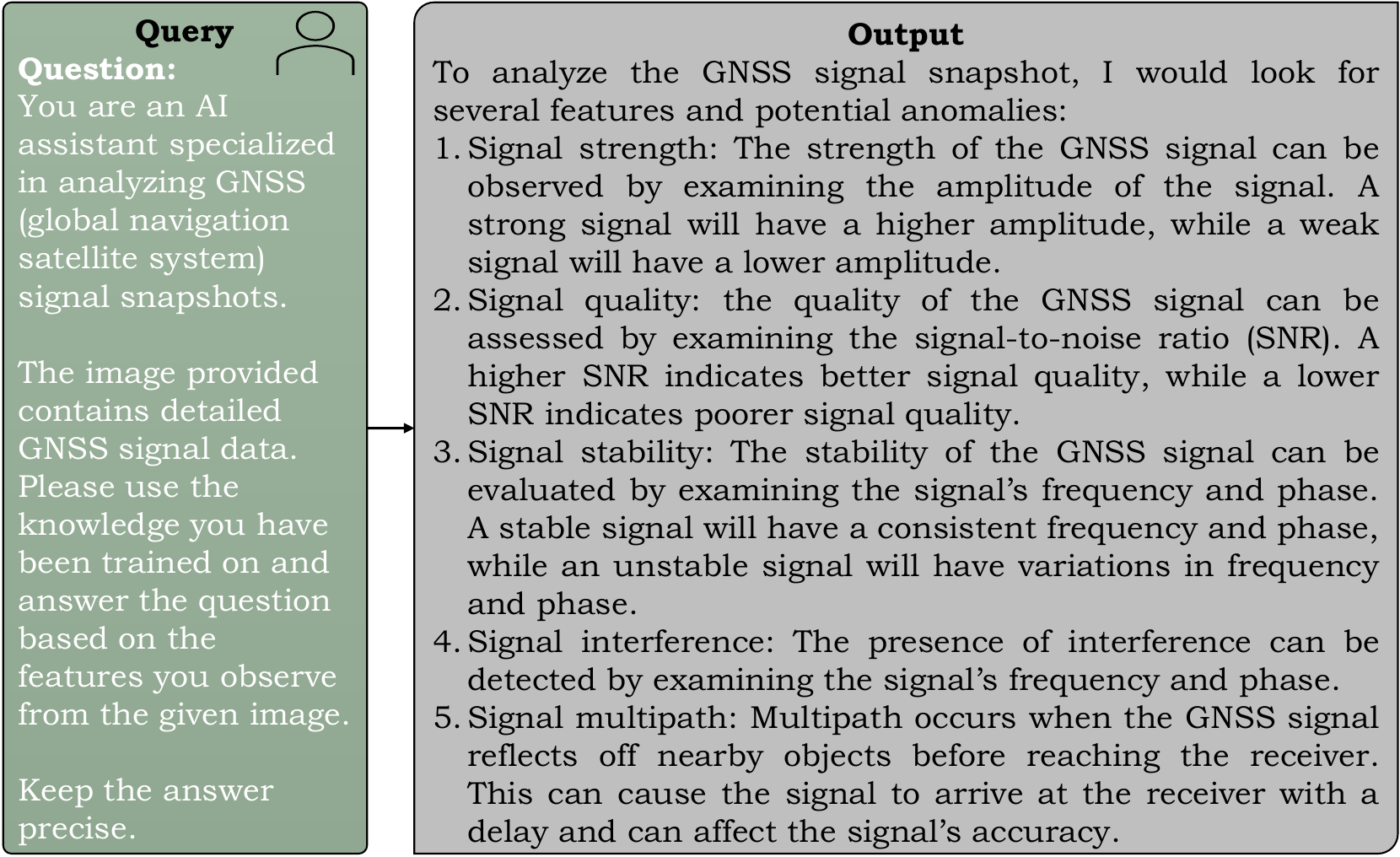}
    \caption{General user query.}
    \label{figure_qa1}
\end{figure}

\begin{figure}[!t]
    \centering
    \includegraphics[width=1.0\linewidth]{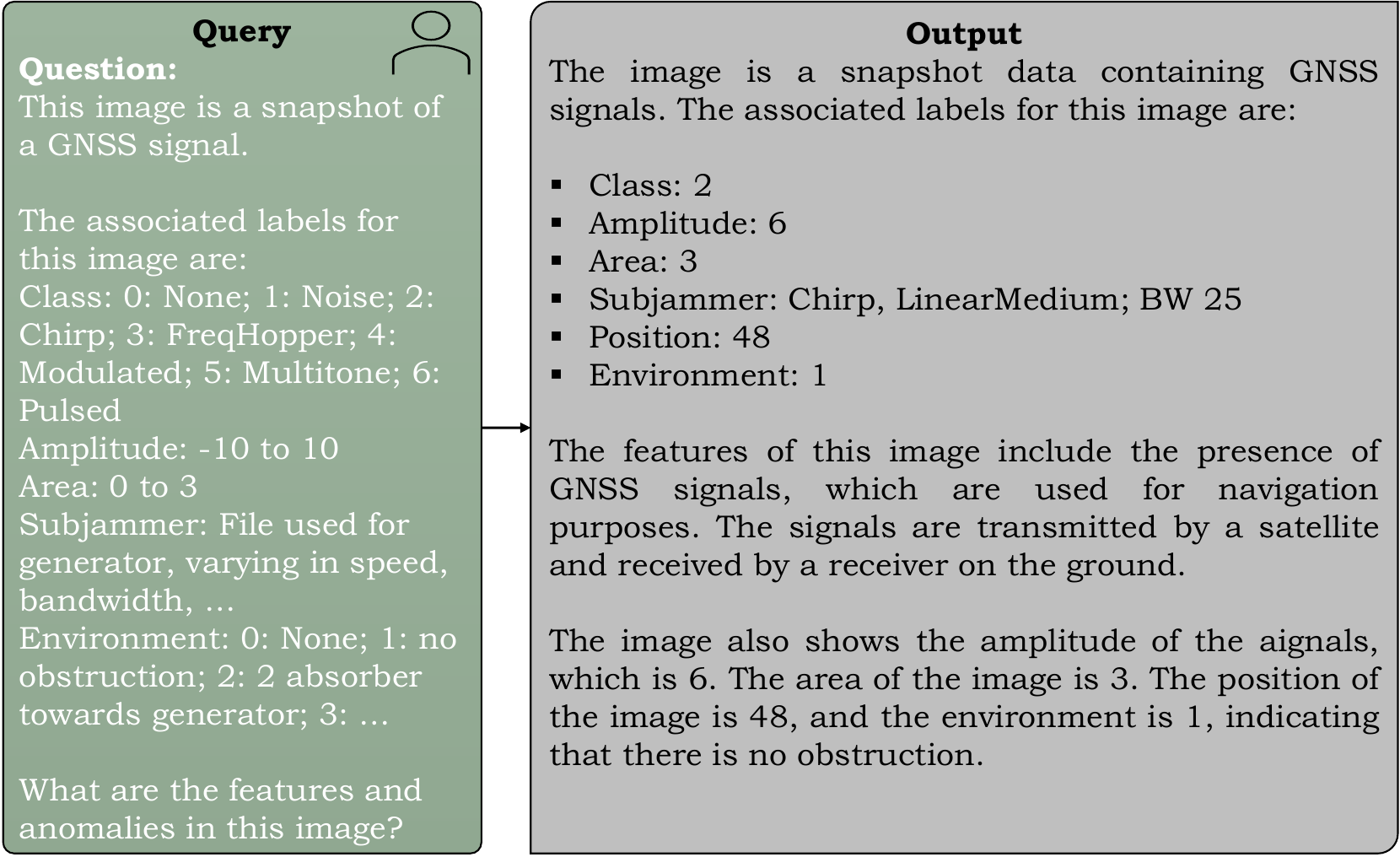}
    \caption{Signal information with general user query.}
    \label{figure_qa2}
\end{figure}

\begin{figure}[!t]
    \centering
    \includegraphics[width=1.0\linewidth]{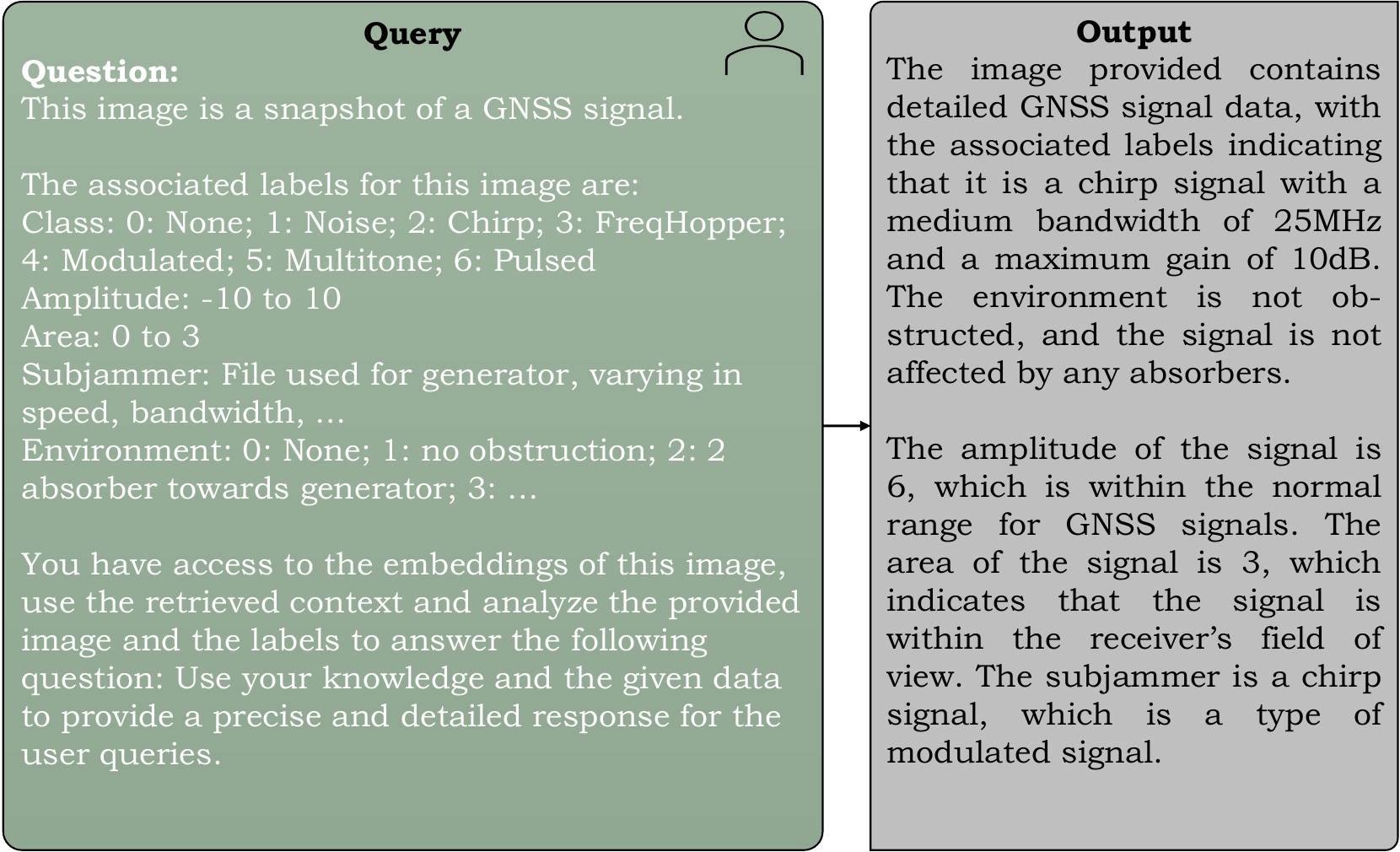}
    \caption{Signal information with detailed user query.}
    \label{figure_qa3}
\end{figure}

\begin{figure}[!t]
    \centering
    \includegraphics[width=1.0\linewidth]{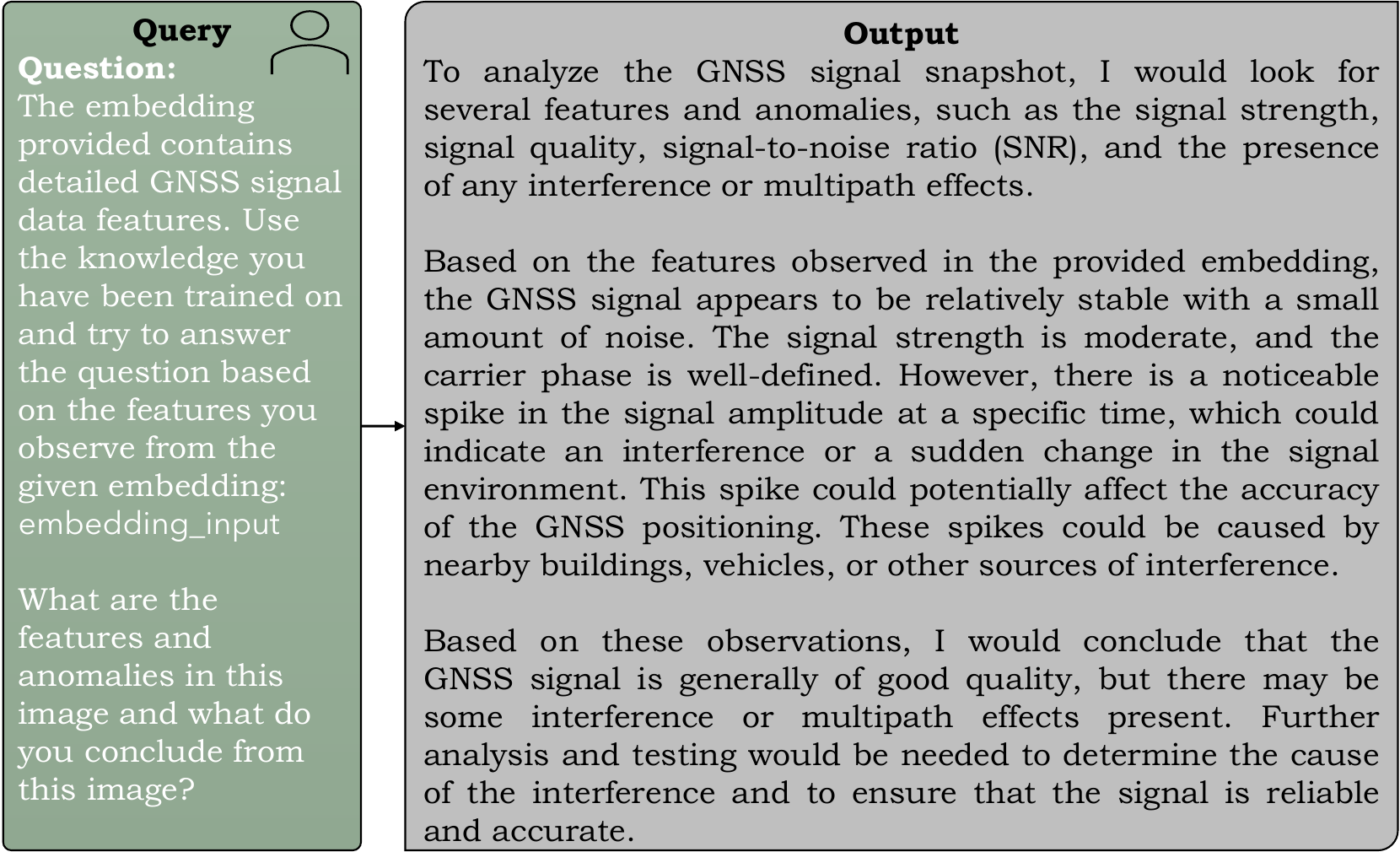}
    \caption{General user query with interpretation.}
    \label{figure_qa4}
\end{figure}

\textbf{Prompt Engineering.} Evaluating predictions of language models is challenging because their outputs are inherently descriptive, even for tasks like classification, making it harder to align predictions with rigid evaluation metrics. In general, the analysis suggests that the model can accurately interpret the image dimensions without requiring additional information about the GNSS signal, as demonstrated by the following response: ``\textit{The size of the snapshot is 32 times 1024, with 32 being the length of the snapshot and 1024 being the number of channels.}'' When an exemplary of \textit{chirp} interference is provided (see Figure~\ref{figure_exmp_samples2}), the model accurately describes the snapshot, stating: ``\textit{The signal appears to be a chirp signal, which is a type of modulated signal used in GNSS systems and are characterized by a linear frequency sweep, which means that the frequency of the signal increases or decreases linearly over time.}'' Figure~\ref{figure_qa1} to \ref{figure_qa4} presents four user query questions with varying levels of detail and their corresponding outputs. In Figure~\ref{figure_qa1}, where the prompt lacks in-context learning, the model's output is general, describing potential features of possible anomalies without interpreting the actual image input. With in-context learning (as shown in Figures~\ref{figure_qa2}, \ref{figure_qa3}, and \ref{figure_qa4}), the model interprets the given image input more accurately and provides detailed description. In Figure~\ref{figure_qa2}, when all possible interference characteristics are provided with an embedding input, the model successfully classifies the interference and offers detailed characteristics, such as amplitude and bandwidth, while accurately interpreting the image without multipath effects. In Figure~\ref{figure_qa3}, the user query is modified to instruct the model to utilize the knowledge of the CLIP model. The resulting output is more descriptive, yet still correctly identifies interference characteristics. However, when all possible label information is removed (as seen in Figure~\ref{figure_qa4}), the model describes the image input and its anomalies, noting, for instance, ``\textit{with a small amount of noise}'' and ``\textit{there is a noticeable spike in the signal amplitude}''.

\textbf{Hyperparameter Evaluation.} We next evaluate the impact of the LLaVA hyperparameters
on the model's output. The \textit{temperature} parameter (ranging from 0 to 1) is employed to modulate the probabilities of the next token, thereby controlling the randomness and creativity of the model’s predictions. As the temperature value decreases, the model's output becomes more deterministic and repetitive. We observe that with values between 0.6 and 1.0, the model generates more diverse and potentially creative outputs. The $\text{top}_{\text{k}}$ parameter, where $1 < \text{top}_{\text{k}} < 100$, allows the model to select randomly from the top k tokens based on their respective probabilities. A $\text{top}_{\text{k}}$ value of 100 results in more explanatory but less accurate responses (as illustrated in Figure~\ref{figure_qa1}). In contrast, a $\text{top}_{\text{k}} = 40$ value of 40 yields a more accurate model output.


\setlength{\intextsep}{6pt}
\setlength{\columnsep}{12pt}
\begin{wrapfigure}{R}{4.2cm}
    \begin{minipage}[b]{1.0\linewidth}
        \centering
        \includegraphics[trim=10 10 10 10, clip, width=1.0\linewidth]{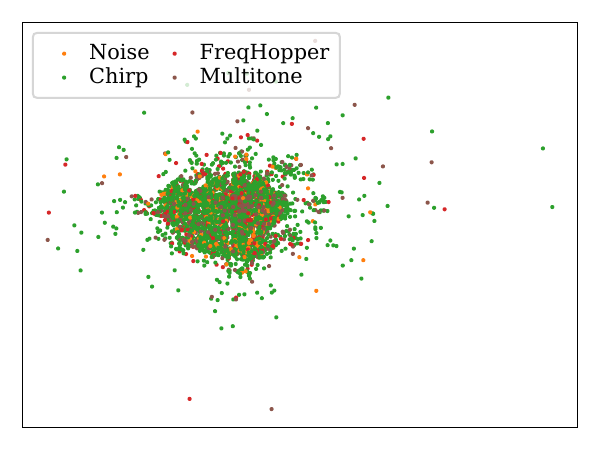}
        \caption{Embeddings of four classes using t-SNE~\cite{maaten_hinton}.}
        \label{figure_embedding}
    \end{minipage}
\end{wrapfigure}

\textbf{Embedding Analysis.} Figure~\ref{figure_embedding} illustrates feature embeddings for four different labels in the vector store of the vision encoder, i.e., the CLIP model, of output size 512. We utilize the t-distributed stochastic neighbor embedding (t-SNE)~\cite{maaten_hinton} with perplexity of 30, an initial momentum of 0.5, and a final momentum of 0.8. We recognize the challenge in distinguishing between the classes due to the absence of distinct clusters, resulting in overlapping scatter plots. Nevertheless, the LLM is still capable of accurately interpreting the embeddings. This observation encourages us to fine-tune the model in future work to achieve improved feature embeddings.

\begin{table}[t!]
\begin{center}
    \vspace{0.3cm}
    \caption{Evaluation results for image classification models trained for 10 epochs. The MSE is utilized as the evaluation metric for both the signal power level and bandwidth.}
    \label{label_results_state_of_the_art}
    \begin{tabular}{ p{1.3cm} | p{0.5cm} | p{0.5cm} | p{0.5cm} | p{0.5cm}}
    \multicolumn{1}{c|}{\textbf{Model}} & \multicolumn{1}{c|}{\textbf{Intf. Type}} & \multicolumn{1}{c|}{\textbf{Subjammer}} & \multicolumn{1}{c|}{\textbf{Power}} & \multicolumn{1}{c}{\textbf{Bandwidth}} \\ \hline
    \multicolumn{1}{l|}{ResNet18} & \multicolumn{1}{r|}{81} & \multicolumn{1}{r|}{47} & \multicolumn{1}{r|}{0.3720} & \multicolumn{1}{r}{0.1993} \\
    \multicolumn{1}{l|}{BEiT} & \multicolumn{1}{r|}{99} & \multicolumn{1}{r|}{80} & \multicolumn{1}{r|}{0.0461} & \multicolumn{1}{r}{0.0217} \\
    \multicolumn{1}{l|}{DeiT} & \multicolumn{1}{r|}{76} & \multicolumn{1}{r|}{26} & \multicolumn{1}{r|}{0.4252} & \multicolumn{1}{r}{0.3948} \\
    \multicolumn{1}{l|}{Swim} & \multicolumn{1}{r|}{99} & \multicolumn{1}{r|}{75} & \multicolumn{1}{r|}{0.1071} & \multicolumn{1}{r}{0.0259} \\
    \multicolumn{1}{l|}{CLIP} & \multicolumn{1}{r|}{81} & \multicolumn{1}{r|}{34} & \multicolumn{1}{r|}{0.3352} & \multicolumn{1}{r}{0.2742} \\
    \multicolumn{1}{l|}{ViT} & \multicolumn{1}{r|}{95} & \multicolumn{1}{r|}{67} & \multicolumn{1}{r|}{0.1630} & \multicolumn{1}{r}{0.0862} \\
    \end{tabular}
\end{center}
\end{table}

\textbf{Benchmark of Vision Models.} Table~\ref{label_results_state_of_the_art} presents a comparison of state-of-the-art vision models on the snapshot dataset for a multi-task problem, encompassing interference classification, subjammer type identification, signal-to-noise ratio estimation, and bandwidth prediction. The models included in the comparison are ResNet18~\cite{heublein_feigl_crpa}, BEiT~\cite{bao_dong_piao}, DeiT~\cite{touvron_cord}, the Swin Transformer~\cite{liu_lin_cao}, CLIP~\cite{radford_kim}, and ViT~\cite{dosovitskiy_beyer}. For GNSS interference detection methods, see \cite{heublein_raichur_ion}. Notably, the vision models BEiT, Swin, and ViT demonstrate significant performance improvements over traditional methods based on ResNet18 across all four tasks. Heublein et al.~\cite{heublein_raichur_ion} achieved an accuracy of 96.15\% on the independent single-task test dataset using a ResNet18 model. In comparison, our LLM surpasses the state-of-the-art, attaining an accuracy of 96.87\%. With the introduction of our proposed method leveraging LLaVa, these results can now be effectively interpreted and analyzed.

\textbf{Time.} The inference time per GNSS snapshot is less than $50\,ms$, making it suitable for real-time applications. However, the primary latency arises during the initial loading phase of the vision encoder and the LLaVA preprocessor. Once loaded, the per-snapshot inference remains highly efficient.
\section{Conclusion}
\label{label_conclusion}

We proposed a pipeline for GNSS interference monitoring, integrating the vision encoder CLIP, the vector store FAISS, and the language model LLaVA. The model's predictions are evaluated based on user queries consisting of multimodal image-text inputs. Specifically, the image input is a GNSS snapshot containing interference, and the accompanying text is a question designed to identify the snapshot's characteristics. Through the evaluation of \textit{task instruction prompting}, we demonstrated that providing detailed contextual information within the query leads to more accurate and comprehensive model outputs. Additionally, by employing \textit{in-context learning}, we showed that a sequence of related examples is essential for the model to generalize effectively from the provided context. In conclusion, our proposed framework enables non-expert users to effectively evaluate GNSS snapshots.

In future work, we will incorporate \textit{retrieval-based prompting}, a method that involves selecting prompts or context using retrieval techniques. In this approach, the model retrieves relevant prompts or context from a prompt pool or an external knowledge base to guide its generation or decision-making process. This can be formally defined as $\mathcal{C} = \mathcal{R}(\mathbf{S}_v, t)$, where $\mathcal{R}$ represents the retrieval method that identifies pertinent prompts or context based on the image $\mathbf{S}_v$ and text $t$ inputs. The external knowledge base may include data from various experts who have manually labeled and described GNSS samples with interferences. The retrieved context $\mathcal{C}$ is subsequently used to guide the model's generation or decision-making process. It is important to note that the retrieval method $\mathcal{R}$ can vary depending on the specific approach and the available prompt pool or knowledge base. This method allows the model to leverage existing information, thereby enhancing its performance by incorporating relevant prompts or context during the generation process.

\clearpage

\bibliography{WCNC2025}
\bibliographystyle{IEEEtran}

\end{document}